\pdfoutput=1
\documentclass[11pt]{article}

\usepackage[]{emnlp2021}

\usepackage{times}
\usepackage{latexsym}

\usepackage[T1]{fontenc}
\usepackage[utf8]{inputenc}

\usepackage{microtype}

\usepackage[T1]{fontenc}
\usepackage{type1cm}
\usepackage[american]{babel}
\usepackage{url}
\usepackage{amssymb}
\usepackage{amsmath}
\usepackage{multirow}
\usepackage{xspace}
\DeclareMathAlphabet{\mathcalbf}{OMS}{pzc}{b}{n}
\usepackage{microtype}
\usepackage{booktabs}

\usepackage[hang,flushmargin]{footmisc}
\newcommand{\Ni}{(1)~}
\newcommand{\Nii}{(2)~}
\newcommand{\Niii}{(3)~}

\usepackage[pdftex]{graphicx}
\usepackage{tabularx}
\usepackage{booktabs}
\usepackage{subcaption}
\DeclareGraphicsExtensions{.pdf,.ai,.jpg,.png}
\setkeys{Gin}{pagebox=artbox}

\graphicspath{{./}}

\newcommand{\bsfigure}[3][]{%
	\begin{figure}[t]
		\centering
		\includegraphics[#1]{#2}
		\caption{#3}\label{#2}%
	\end{figure}
}

\RequirePackage{type1cm}% CM-Fonts including Math-Fonts in all sizes.
\RequirePackage{color}
\RequirePackage{soul}
\setstcolor{blue}
\definecolor{lightgray}{rgb}{0.95,0.95,0.95}
\definecolor{lightgreen}{rgb}{0.56,0.93,0.56}
\definecolor{lightblue}{rgb}{0.3,0.3,0.9}
\definecolor{tgray}{rgb}{0.5,0.5,0.5}

\newsavebox\bscombox
\newcommand{\bscom}[3][]{%
	\sbox{\bscombox}{\fontsize{8}{9}\selectfont#1#2#3}
	\noindent
	\st{#2}{\selectfont
		\color{blue}#3\ifx\\#1\\\else{\fontsize{8}{9}\selectfont\color{violet}[#1]}\fi
	}
}

\begin{document}

	% If the title and author information does not fit in the area allocated, uncomment the following
%
%\setlength\titlebox{<dim>}
%
% and set <dim> to something 5cm or larger.

\title{Key Point Analysis via Contrastive Learning \\ and Extractive Argument Summarization}

% Author information can be set in various styles:
% For several authors from the same institution:
% \author{Author 1 \and ... \and Author n \\
%         Address line \\ ... \\ Address line}
% if the names do not fit well on one line use
%         Author 1 \\ {\bf Author 2} \\ ... \\ {\bf Author n} \\
% For authors from different institutions:
% \author{Author 1 \\ Address line \\  ... \\ Address line
%         \And  ... \And
%         Author n \\ Address line \\ ... \\ Address line}
% To start a seperate ``row'' of authors use \AND, as in
% \author{Author 1 \\ Address line \\  ... \\ Address line
%         \AND
%         Author 2 \\ Address line \\ ... \\ Address line \Andc
%         Author 3 \\ Address line \\ ... \\ Address line}

\newcommand{\lei}{\textsuperscript{$\ddagger$}}
\newcommand{\pb}{\textsuperscript{$\dagger$}}
\newcommand{\bif}{\textsuperscript{$\star$}}

\author{%
	Milad Alshomary \pb
	\quad Timon Gurcke \pb
	\quad Shahbaz Syed \lei
	\quad Philipp Heinisch \bif \\[1.5ex]
	\bfseries Maximilian Spliethöver \pb \hspace{1.5ex}
	\bfseries Philipp Cimiano \bif \hspace{1.5ex}
	\bfseries Martin Potthast \lei \hspace{1.5ex}
	\bfseries Henning Wachsmuth \pb \\[1.5ex] 
	\pb Paderborn University, Germany,   {\tt milad.alshomary@upb.de}\\
	\lei Leipzig University, Germany,  {\tt <first>.<last>@uni-leipzig.de} \\
	\bif Bielefeld University, Germany,  {\tt <first>.<last>@uni-bielefeld.de  } \\
}

\maketitle

\begin{abstract}
Key point analysis is the task of extracting a set of concise and high-level statements from a given collection of arguments, representing the gist of these arguments. This paper presents our proposed approach to the Key Point Analysis shared task, collocated with the 8th Workshop on Argument Mining. The approach integrates two complementary components. One component employs contrastive learning via a siamese neural network for matching arguments to key points; the other is a graph-based extractive summarization model for generating key points. In both automatic and manual evaluation, our approach was ranked best among all submissions to the shared task.
\end{abstract}

	\section{Introduction}
\label{sec:introduction}

Informed decision-making on a controversial issue usually requires considering several pro and con arguments. To answer the question ``Is organic food healthier?'', for example, people may query a search engine that retrieves arguments from diverse sources such as news editorials, debate portals, and social media discussions, which can then be compared and weighed. However, given the constant stream of digital information, this process may be time-intensive and overwhelming. Search engines and similar support systems may therefore benefit from employing argument summarization, that is, the generated summaries may aid the decision-making by helping users quickly choose relevant arguments with a specific stance towards the topic.

Argument summarization has been tackled both for single documents \cite{syed:2020} and multiple documents \cite{bhatia:2014,egan:2016}. A specific multi-document scenario introduced by \newcite{bar-haim:2020a} is \emph{key point analysis} where the goal is to map a collection of arguments to a set of salient key points (say, high-level arguments) to provide a quantitative summary of these arguments.

The Key Point Analysis (KPA) shared task by \newcite{roni:2021}%
\footnote{\url{https://2021.argmining.org/shared_task_ibm}, last accessed: 2021-08-08}
invited systems for two complementary subtasks: {\em matching arguments to key points} and {\em generating key points from a given set of arguments} (Section \ref{task}). As part of this shared task, we present an approach with two complementary components, one for each subtask. For key point matching, we propose a model that learns a semantic embedding space where instances that match are closer to each other while non-matching instances are further away from each other. We learn to embed instances by utilizing a contrastive loss function in a siamese neural network \cite{bromley:1994}. For the key point generation, we present a graph-based extractive summarization approach similar to the work of \newcite{alshomary:2020a}. It utilizes a PageRank variant to rank sentences in the input arguments by quality and predicts the top-ranked sentences to be key points. In an additional experiment, we also investigated an approach that performs aspect identification on arguments, followed by aspect clustering to ensure diversity. Finally, arguments with the best coverage of these diverse aspects are extracted as key points.

Our approaches yielded the top performance among all submissions to the shared task in both quantitative and qualitative evaluation conducted by the organizers of the shared task (Section \ref{experiments-and-evaluation}).\footnote{The code is available under \url{https://github.com/webis-de/ArgMining-21}}.
	\section{Related Work}
\label{sec:relatedwork}

In summarization, arguments are relatively understudied compared to other document types such as news articles or scientific literature, but a few approaches have come up in the last years.

In an extractive manner, argument mining has been employed to identify the main claim as the summary of an argument~\cite{petasis:2016,daxenberger:2017}. \citet{wang:2016} used a sequence-to-sequence model for the abstractive summarization of arguments from online debate portals. A complementary task of generating conclusions as informative argument summaries was introduced by \newcite{syed:2021}. Similar to \newcite{alshomary:2020b} who inferred a conclusion's target with a triplet neural network, we rely on contrastive learning here, using a siamese network though. Also, we build upon ideas of \newcite{alshomary:2020a} who proposed a graph-based model using PageRank~\cite{page:1999} that extracts the argument's conclusion and the main supporting reason as an extractive summary. All these works represent the single-document summarization paradigm where only one argument is summarized at a time, whereas the given shared task is a multi-document summarization setting.

The first approaches to multi-document argument summarization aimed to identify the main points of online discussions. Among these, \citet{egan:2016} grouped verb frames into pattern clusters that serve as input to a structured summarization pipeline, whereas \citet{Misra:2016} proposed a more condensed approach by directly extracting argumentative sentences, summarized by similarity clustering. \newcite{bar-haim:2020a} continued this line of research by introducing the notion of \emph{key points} and contributing the ArgsKP corpus, a collection of arguments mapped to manually-created key points. These key points are concise and self-contained sentences that capture the gist of the arguments. Later, \newcite{bar-haim:2020b} proposed a quantitative argument summarization framework that automatically extracts key points from a set of arguments. Building upon this research, our approach aims to increase the quality of such generated key points, including a strong relation identifier between arguments and key points. 
	\section{Task Description}
\label{task}

In the context of computational argumentation, \newcite{bar-haim:2020a} introduced the notion of a {\em key point} as a high-level argument that resembles a natural language summary of a collection of more descriptive arguments. Specifically, the authors defined a good key point as being ``general enough to match a significant portion of the arguments, yet informative enough to make a useful summary.'' In this context, the KPA shared task consists of two subtasks as described below:
\begin{enumerate}
\setlength{\itemsep}{0pt}
\item 
\textit{Key point matching.} Given a set of arguments on a certain topic that are grouped by their stance and a set of key points, assign each argument to a key point.
\item 
\textit{Key point generation and matching.} Given a set of arguments on a certain topic that are grouped by their stance, first generate five to ten key points summarizing the arguments. Then, match each argument in the set to the generated key points (as in the previous track).
\end{enumerate}

\paragraph{Data}

We start from the dataset provided by the organizers as described in \newcite{roni:2021}. The dataset contains 28~controversial topics, with 6515~arguments and a total of 243 key points. For each argument, its stance towards the topic as well as a quality score are given. Each topic is represented by at least three key points, with at least one key point per stance and at least three arguments matched to a key point. From the given arguments, 4.7\%~are unmatched, 67.5\%~belong to a single key point, and 5.0\%~belong to multiple key points. The remaining 22.8\%~of the arguments have ambiguous labels, meaning that the annotators could not agree on a correct matching to the key points. The final dataset contains 24,093~argument-key point pairs, of which 20.7\%~are labeled as matching. To develop our approach, we use the split as provided by the organizers with 24~topics for training, four topics for validation, and three topics for testing.

% Ambigious labels should be described in the evaluation part

% topic stance key-point <-- argument
% ~24K argument/key-point pairs
% 28 controversial topics
% show examples
% statistics

% s (argument, key point)
% pairs with binary labels indicating whether the argument is matched to the key point.

% each key-point ahs at least 3 arguments
% key-points created by a human

% The final dataset has 24,093 labeled (argument,
% key point) pairs, of which 4,998 pairs (20.7%) are
% positive. It has 6,515 arguments (232.67 per topic),
% and 243 key points (8.67 key points per topic). For
% each pair, the dataset also specifies the topic and
% the stance of the argument towards the topic.

% what is train val test split

% quality score for each argument

% Each fold comprises 3 test topics,
% 17 train topics and 4 development topics.

% (TBD) Description of the data provided by the organizers

% Full:
% 243
% 6515
% 28
% Train:
% 207
% 5583
% 24
% Dev:
% 36
% 932
% 4
% Test:
% 33
% 723
% 3
	\section{Approach}
\label{sec:approach}

Our approach consists of two components, each corresponding to one subtask of the KPA shared task. The first subtask of matching arguments to key points is modeled as a contrastive learning task using a siamese neural network. The second subtask requires generating key points for a collection of arguments and then matching them to the arguments. We investigated two models for this subtask: One is a graph-based extractive summarization model utilizing PageRank \cite{page:1999} to extract sentences representing the key points; the other identifies aspects from the arguments and selects the most representative sentences that maximize the coverage of these aspects as the key points.

\subsection{Key Point Matching}
\label{key-point-matching}

Conceptually, we consider pairs of arguments and key points that are close to each other in a semantic embedding space as possible candidates for matching. Furthermore, we seek to transform this space into a new embedding space where matching pairs are closer and the non-matching ones are more distant from each other (Figure \ref{key-point-argument-mapping.ai}). To do so, we utilize a siamese neural network with a contrastive loss function.

Specifically, in the training phase, the input is a topic along with a key point, an argument, and a label (matching or not). First, we use a pretrained language model to encode the tokens of the argument as well as those of the concatenation of the topic and the key point. Then, we pass their embeddings through a siamese neural network, which is a mean-pooling layer that aggregates the token embeddings of each input, resulting in two sentence-level embeddings. We compute the contrastive loss using these embeddings as follows:
\begin{equation*}
\mathcal{L} \;\;=\;\; -y \cdot \log(\hat{y}) \;+\; (1-y) \cdot \log(1-\hat{y})
\end{equation*}
where $\hat{y}$ is the cosine similarity of the embeddings, and $y$ reflects whether a pair matches (1) or not (0).

\bsfigure{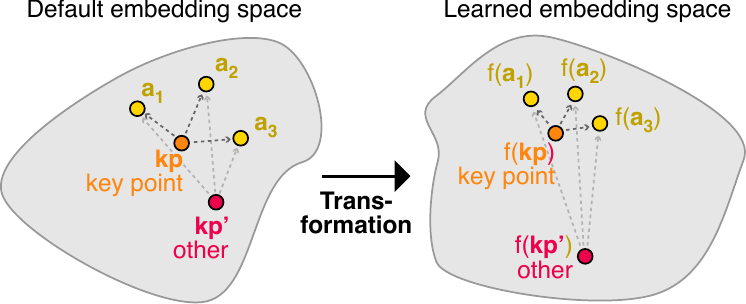}{We learn to transform an embedding space into a new space in which matching pairs of key point and argument (e.g., $kp$ and $a_1$) are closer to each other, and the distance between non-matching pairs (e.g., $kp'$ and $a_1$) is larger. For simplicity, $kp$ and $kp'$ each represent a concatenation of key point and topic.}

\subsection{Key Point Generation}

Our primary model for key point generation is a graph-based extractive summarization model. Additionally, we also investigate clustering the aspects of the given collection of arguments.

\paragraph{Graph-based Summarization} 

Following the work of \newcite{alshomary:2020a},  we first construct an undirected graph with the arguments' sentences as nodes. As a filtering step, we compute argument quality scores for each sentence as \newcite{toledo:2019} and exclude low-quality arguments from the graph. Next, we employ our key point matching model (Section \ref{key-point-matching}) to compute the edge weight between two nodes as the pairwise matching score of the corresponding sentences. Only nodes with a score above a defined threshold are connected via an edge. An example graph is sketched in Figure~\ref{key-point-pagerank.ai}. Finally, we use a variant of PageRank \cite{page:1999} to compute an importance score $P(s_i)$ for each sentence $ s_i $ as follows:
\begin{equation}\label{equation-key-point-generation}
\small
\begin{aligned}
P(s_i) \; = {} \; (1-d) & \cdot \sum_{s_j \not = s_i}^{} \frac{match(s_i, s_j)}{\sum_{s_k \not = s_j} match(s_j, s_k)} P(s_j) \; \\
+ \; d \,\,& \cdot \frac{qual(s_i)}{\sum_{s_k}^{} qual(s_k)}
\end{aligned}
\end{equation}
where $d$ is a damping factor that can be configured to bias the algorithm towards the argument quality score $qual$ or the matching score $match$. To ensure diversity, we iterate through the ranked list of sentences (in descending order), adding a sentence to the final set of key points if its maximum matching score with the already selected candidates is below a certain threshold. 

\bsfigure[scale=1]{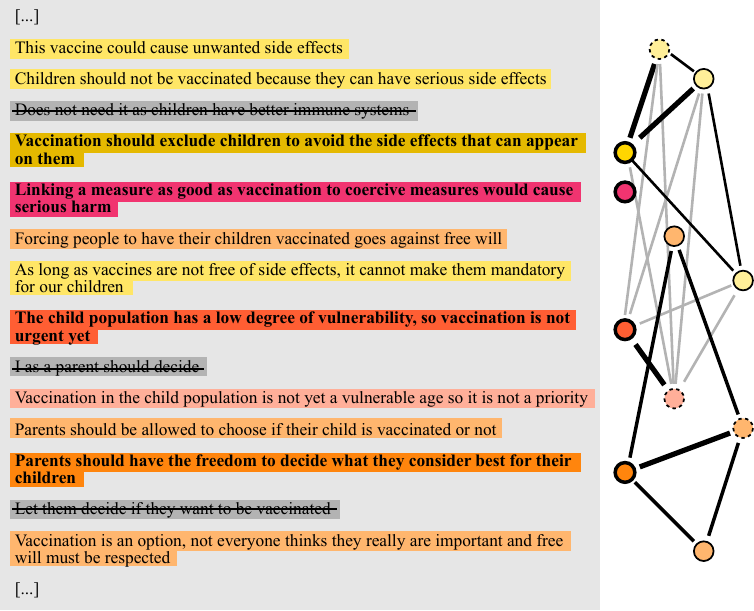} {Example graph of our key point generation approach. Nodes with high saturation are considered to be key points (bold text). Nodes with dashed lines have lower argument quality. Edge thickness represents similarity between two nodes. Notice that the shown arguments do not reflect the view of the authors.}

\paragraph{Aspect Clustering} 

Extracting key points is conceptually similar to identifying aspects~\citep{bar-haim:2020a}, which inspired our clustering approach that selects representative  sentences from multiple aspect clusters as the final key points. 
We employ the tagger of \newcite{schiller:2021} to extract the arguments' aspects (on average, 2.1 aspects per argument).  To tackle the lack of diversity, we follow \newcite{Heinisch:2021} and create $k$ diverse aspect clusters by projecting the extracted aspect phrases to an embedding space. Next, we model the candidate selection of argument sentences as the set cover problem. Specifically, the final set of key points summarizing the arguments for a given topic and stance maximizes the coverage of the set of arguments' aspects. To this end, we apply greedy approximation for selecting our candidates, where an argument sentence is chosen if it covers the maximum number of unique aspect clusters while having the smallest overlap with the clusters covered by the already selected candidates. Also, to avoid redundant key points, we compute its semantic similarity to the already selected candidates in each candidate selection step, and we add it to the final set if its score is below a certain threshold.

	\section{Experiments and Evaluation}
\label{experiments-and-evaluation}

In the following, we present implementation details of our two components, and we report on their quantitative and qualitative results.

\subsection{Key Point Matching}

We employed RoBERTa-large \cite{liu:2019} for encoding the tokens of the two inputs of key point matching to the siamese neural network, which acts as a mean-pooling layer and projects the encoder outputs (matrix of token embeddings) into a sentence embedding of size 768. We used Sentence-BERT \cite{reimers:2019b} to train our model for 10 epochs, with batch size 32, and maximum input length of 70, leaving all other parameters to their defaults. 

For automatic evaluation, we computed both strict and relaxed mean Average Precision (mAP) following \newcite{roni:2021}. In cases where there is no majority label for matching, the relaxed mAP considers them to be a match while the strict mAP considers them as not matching. In the development phase, we trained our model on the training split and evaluated on the validation split provided by the organizers. The strict and relaxed mAP on the validation set were 0.84 and 0.96 respectively. For the final submission, we did a five-fold cross validation on the combined data (training and validation splits) creating an ensemble for the matching (as per the mean score).

\begin{table}[t!]%
\centering%
\small
\renewcommand{\arraystretch}{1}
\setlength{\tabcolsep}{6pt}%
\begin{tabular}{lccc}
\toprule
\bf Approach    & \bf R-1 & \bf R-2  & \bf R-L \\
\midrule
Graph-based Summarization     &  \bf 19.8 &  3.5  &  \bf 18.0\\
Aspect Clustering &  18.9 &  \bf 4.7  &  17.1\\
\bottomrule
\end{tabular}%
\caption{ROUGE scores on the test set for our two approaches to key point generation.}
\label{table-internal-automatic-eval}%
\end{table}

\subsection{Key Point Generation}

For the \textbf{graph-based summarization model}, we employed Spacy \cite{honnibal:2020} to split the arguments into sentences. Similar to \cite{bar-haim:2020b}, only sentences with a minimum of 5 and a maximum of 20 tokens, and not starting with a pronoun, were used for building the graph. Argument quality scores for each sentence were obtained from \text{Project Debater}'s API \cite{toledo:2019}\footnote{Available under: \url{https://early-access-program.debater.res.ibm.com/}}. We selected the thresholds for the parameters \emph{d}, \emph{qual} and \emph{match} in Equation~\ref{equation-key-point-generation} as 0.2, 0.8 and 0.4 respectively, optimizing for ROUGE \cite{lin:2004}. In particular, we computed ROUGE-L between the ground-truth key points and the top 10 ranked sentences as our predictions, averaged over all the topic and stance combinations in the training split. We excluded sentences with a matching score higher than 0.8 with the selected candidates to minimize redundancy.

\begin{table*}[t]
\small
\centering
\renewcommand{\arraystretch}{1.2}
\begin{tabular}{@{}p{0.15\linewidth}cp{0.35\linewidth}p{0.35\linewidth}@{}}
\toprule
\bfseries{Topic} &\bfseries{Stance} & \bfseries{Graph-based Summarization} & \bfseries{Aspect Clustering}\\
\midrule                                                                                        
Routine child vaccinations should be mandatory & Pro  & 
\Ni Child vaccinations should be mandatory to provide decent health care to all.  
\Nii Vaccines help children grow up healthy and avoid dangerous diseases.
\Niii Child vaccinations should be mandatory so our children will be safe and protected.

& 
\Ni Child vaccination is needed for children, they get sick too. 
\Nii Routine child vaccinations should be mandatory to prevent the disease.
\Niii Yes as they protect children from life threatening and highly infectious diseases.
\\
Routine child vaccinations should be mandatory & Con  & 
\Ni Vaccination should exclude children  to avoid the side effects that can appear on them.  
\Nii Parents should have the freedom to decide what they consider best for their children.
\Niii The child population has a low degree of vulnerability, so vaccination is not urgent yet.

& 
\Ni Child vaccination shouldn't be mandatory because the virus isn't effective in children. 
\Nii Child vaccinations should not be mandatory because vaccines are expensive.
\Niii It has not been 100\% proven if the vaccine is effective.
\\
\bottomrule
\end{tabular}
\caption{Examples of keypoints from our proposed approaches. Only the top three key points are shown for brevity. }
\label{table-keypoints-examples}
\end{table*}

For \textbf{aspect clustering}, we created 15 clusters per topic and stance combination. After greedy approximation of the candidate sentences, we removed redundant ones using a threshold of 0.65 for the normalized BERTScore~\cite{Zhang:2020} with the previously selected candidates.

\paragraph{Comparison of both approaches}

To select our primary approach for key point generation,  we first performed an automatic evaluation of the aforementioned models on the test set using ROUGE (Table \ref{table-internal-automatic-eval}). Additionally, we performed a manual evaluation via pairwise comparison of the extracted key points for both models for a given topic and stance. 

Examples of key points from both the models are shown in Table~\ref{table-keypoints-examples}. The key points from graph-based summarization model are relatively longer. This also improves their informativeness, matching findings of \newcite{syed:2021}. For the aspect clustering, we observe that the key points are more focused on specific aspects such as ``disease'' (for Pro) and ``effectiveness'' (for Con). In a real-world application, this may provide the flexibility to choose key points by aspects of interest to the end-user, especially with further improvement of aspect tagger by avoiding non-essential extracted phrases as ``mandatory''. Hence, given the task of generating a quantitative summary of a collection of arguments, we believe that the graph-based summary provides a more comprehensive overview and chose this as our preferred approach for key point generation. 

\subsection{Shared Task's Evaluation Results} 

\begin{table}[t!]%
\centering%
\small
\renewcommand{\arraystretch}{1.1}
\setlength{\tabcolsep}{6pt}%
\begin{tabular}{lcccc}
\toprule
    & \bf KP Matching & \multicolumn{3}{c}{\bf KP Generation}  \\
\cmidrule(l@{5pt}r@{5pt}){2-2} \cmidrule(l@{5pt}r@{5pt}){3-5} 
\bf Approach & \bf mAP/Rank & \bf Rel. & \bf Rep. & \bf Pol. \\
\midrule
\bf bar\_h      & 0.885/1 & 2 & 1 & 1 \\
\bf mspl (ours)  & 0.818/2 & 2 & 1 & 2 \\ 
\bf sohanpat     & 0.491/3 & 4 & 4 & 2 \\ 
\bf peratham     & 0.443/4 & 1 & 3 & 4 \\

\bottomrule
\end{tabular}%
\caption{Final evaluation results of both tracks, comparing our approach (mspl) to the top two submitted approaches, along with  \newcite{bar-haim:2020b} approach (bar\_h). The generated key points were ranked in terms of how relevant (Rel.) and representative (Rep.) of the input arguments, as well as their polarity (Pol.)}
\label{table-final-eval}%
\end{table}

In key point matching, our approach obtained a strict mAP of 0.789 and a relaxed mAP of 0.927 on the test set, the best result among all participating approaches. For the second track, in addition to evaluating the key point matching task, the shared task organizers manually evaluated the generated key points through a crowdsourcing study in which submitted approaches were ranked according to the quality of their generated key points. Table \ref{table-final-eval} presents the evaluation results of the top three submitted approaches, along with the reference approach of \newcite{bar-haim:2020b}. Among the submitted approaches, our approach was ranked the best in both the key point generation task as well as the key point matching task. For complete details on the evaluation, we refer to the task organizers' report \cite{roni:2021}.

	\section{Conclusion}

This paper has presented a framework to tackle the key point analysis of arguments. For matching arguments to key points, we achieved the best performance in the KPA shared task via contrastive learning. For key point generation, we developed a graph-based extractive summarization model that output informative key points of high quality for a collection of arguments. We see abstractive key point generation as part of our future work.

	\bibliography{argmining21-kpa-sharedtask-lit}

\begin{thebibliography}{23}
\expandafter\ifx\csname natexlab\endcsname\relax\def\natexlab#1{#1}\fi

\bibitem[{Alshomary et~al.(2020{\natexlab{a}})Alshomary, D{\"{u}}sterhus, and
  Wachsmuth}]{alshomary:2020a}
Milad Alshomary, Nick D{\"{u}}sterhus, and Henning Wachsmuth.
  2020{\natexlab{a}}.
\newblock \href {https://doi.org/10.1145/3397271.3401186} {Extractive snippet
  generation for arguments}.
\newblock In \emph{Proceedings of the 43rd International {ACM} {SIGIR}
  conference on research and development in Information Retrieval, {SIGIR}
  2020, Virtual Event, China, July 25-30, 2020}, pages 1969--1972. {ACM}.

\bibitem[{Alshomary et~al.(2020{\natexlab{b}})Alshomary, Syed, Potthast, and
  Wachsmuth}]{alshomary:2020b}
Milad Alshomary, Shahbaz Syed, Martin Potthast, and Henning Wachsmuth.
  2020{\natexlab{b}}.
\newblock \href {https://doi.org/10.18653/v1/2020.acl-main.399} {Target
  inference in argument conclusion generation}.
\newblock In \emph{Proceedings of the 58th Annual Meeting of the Association
  for Computational Linguistics, {ACL} 2020, Online, July 5-10, 2020}, pages
  4334--4345. Association for Computational Linguistics.

\bibitem[{Bar-Haim et~al.(2020{\natexlab{a}})Bar-Haim, Eden, Friedman, Kantor,
  Lahav, and Slonim}]{bar-haim:2020a}
Roy Bar-Haim, Lilach Eden, Roni Friedman, Yoav Kantor, Dan Lahav, and Noam
  Slonim. 2020{\natexlab{a}}.
\newblock \href {https://www.aclweb.org/anthology/2020.acl-main.371} {From
  arguments to key points: {T}owards automatic argument summarization}.
\newblock In \emph{Proceedings of the 58th Annual Meeting of the Association
  for Computational Linguistics}, pages 4029--4039. Association for
  Computational Linguistics.

\bibitem[{Bar-Haim et~al.(2020{\natexlab{b}})Bar-Haim, Kantor, Eden, Friedman,
  Lahav, and Slonim}]{bar-haim:2020b}
Roy Bar-Haim, Yoav Kantor, Lilach Eden, Roni Friedman, Dan Lahav, and Noam
  Slonim. 2020{\natexlab{b}}.
\newblock \href {https://doi.org/10.18653/v1/2020.emnlp-main.3} {Quantitative
  argument summarization and beyond: Cross-domain key point analysis}.
\newblock In \emph{Proceedings of the 2020 Conference on Empirical Methods in
  Natural Language Processing, {EMNLP} 2020, Online, November 16-20, 2020},
  pages 39--49. Association for Computational Linguistics.

\bibitem[{Bhatia et~al.(2014)Bhatia, Biyani, and Mitra}]{bhatia:2014}
Sumit Bhatia, Prakhar Biyani, and Prasenjit Mitra. 2014.
\newblock \href {https://doi.org/10.3115/v1/d14-1226} {Summarizing online forum
  discussions - can dialog acts of individual messages help?}
\newblock In \emph{Proceedings of the 2014 Conference on Empirical Methods in
  Natural Language Processing, {EMNLP} 2014, October 25-29, 2014, Doha, Qatar,
  {A} meeting of SIGDAT, a Special Interest Group of the {ACL}}, pages
  2127--2131. {ACL}.

\bibitem[{Bromley et~al.(1994)Bromley, Guyon, LeCun, S{\"a}ckinger, and
  Shah}]{bromley:1994}
Jane Bromley, Isabelle Guyon, Yann LeCun, Eduard S{\"a}ckinger, and Roopak
  Shah. 1994.
\newblock Signature verification using a ``siamese'' time delay neural network.
\newblock In \emph{Advances in neural information processing systems}, pages
  737--744.

\bibitem[{Daxenberger et~al.(2017)Daxenberger, Eger, Habernal, Stab, and
  Gurevych}]{daxenberger:2017}
Johannes Daxenberger, Steffen Eger, Ivan Habernal, Christian Stab, and Iryna
  Gurevych. 2017.
\newblock \href {https://doi.org/10.18653/v1/D17-1218} {What is the essence of
  a claim? cross-domain claim identification}.
\newblock In \emph{Proceedings of the 2017 Conference on Empirical Methods in
  Natural Language Processing}, pages 2055--2066, Copenhagen, Denmark.
  Association for Computational Linguistics.

\bibitem[{Egan et~al.(2016)Egan, Siddharthan, and Wyner}]{egan:2016}
Charlie Egan, Advaith Siddharthan, and Adam~Z. Wyner. 2016.
\newblock \href {https://doi.org/10.18653/v1/w16-2816} {Summarising the points
  made in online political debates}.
\newblock In \emph{Proceedings of the Third Workshop on Argument Mining, hosted
  by the 54th Annual Meeting of the Association for Computational Linguistics,
  ArgMining@ACL 2016, August 12, Berlin, Germany}. The Association for Computer
  Linguistics.

\bibitem[{Friedman et~al.(2021)Friedman, Dankin, Katz, Hou, and
  Slonim}]{roni:2021}
Roni Friedman, Lena Dankin, Yoav Katz, Yufang Hou, and Noam Slonim. 2021.
\newblock Overview of {KPA}-2021 shared task: Key point based quantitative
  summarization.

\bibitem[{Heinisch and Cimiano(2021)}]{Heinisch:2021}
Philipp Heinisch and Philipp Cimiano. 2021.
\newblock \href {https://doi.org/doi:10.1515/itit-2020-0054} {A multi-task
  approach to argument frame classification at variable granularity levels}.
\newblock \emph{it - Information Technology}, 63(1):59--72.

\bibitem[{Honnibal et~al.(2020)Honnibal, Montani, Van~Landeghem, and
  Boyd}]{honnibal:2020}
Matthew Honnibal, Ines Montani, Sofie Van~Landeghem, and Adriane Boyd. 2020.
\newblock \href {https://doi.org/10.5281/zenodo.1212303} {{spaCy:
  Industrial-strength Natural Language Processing in Python}}.

\bibitem[{Lin(2004)}]{lin:2004}
Chin-Yew Lin. 2004.
\newblock \href {https://www.aclweb.org/anthology/W04-1013} {{ROUGE}: A package
  for automatic evaluation of summaries}.
\newblock In \emph{Text Summarization Branches Out}, pages 74--81, Barcelona,
  Spain. Association for Computational Linguistics.

\bibitem[{Liu and Lapata(2019)}]{liu:2019}
Yang Liu and Mirella Lapata. 2019.
\newblock \href {https://doi.org/10.18653/v1/D19-1387} {Text summarization with
  pretrained encoders}.
\newblock In \emph{Proceedings of the 2019 Conference on Empirical Methods in
  Natural Language Processing and the 9th International Joint Conference on
  Natural Language Processing (EMNLP-IJCNLP)}, pages 3730--3740, Hong Kong,
  China. Association for Computational Linguistics.

\bibitem[{Misra et~al.(2016)Misra, Ecker, and Walker}]{Misra:2016}
Amita Misra, Brian Ecker, and Marilyn Walker. 2016.
\newblock \href {https://doi.org/10.18653/v1/W16-3636} {Measuring the
  similarity of sentential arguments in dialogue}.
\newblock In \emph{Proceedings of the 17th Annual Meeting of the Special
  Interest Group on Discourse and Dialogue}, pages 276--287, Los Angeles.
  Association for Computational Linguistics.

\bibitem[{Page et~al.(1999)Page, Brin, Motwani, and Winograd}]{page:1999}
Lawrence Page, Sergey Brin, Rajeev Motwani, and Terry Winograd. 1999.
\newblock The page{R}ank citation ranking: {B}ringing order to the web.
\newblock Technical report, Stanford InfoLab.

\bibitem[{Petasis and Karkaletsis(2016)}]{petasis:2016}
Georgios Petasis and Vangelis Karkaletsis. 2016.
\newblock \href {https://www.aclweb.org/anthology/W16-2811/} {Identifying
  argument components through textrank}.
\newblock In \emph{Proceedings of the Third Workshop on Argument Mining, hosted
  by the 54th Annual Meeting of the Association for Computational Linguistics,
  ArgMining@ACL 2016, August 12, Berlin, Germany}.

\bibitem[{Reimers and Gurevych(2019)}]{reimers:2019b}
Nils Reimers and Iryna Gurevych. 2019.
\newblock \href {http://arxiv.org/abs/1908.10084} {Sentence-bert: Sentence
  embeddings using siamese bert-networks}.
\newblock In \emph{Proceedings of the 2019 Conference on Empirical Methods in
  Natural Language Processing}. Association for Computational Linguistics.

\bibitem[{Schiller et~al.(2021)Schiller, Daxenberger, and
  Gurevych}]{schiller:2021}
Benjamin Schiller, Johannes Daxenberger, and Iryna Gurevych. 2021.
\newblock \href {https://www.aclweb.org/anthology/2021.naacl-main.34/}
  {Aspect-controlled neural argument generation}.
\newblock In \emph{Proceedings of the 2021 Conference of the North American
  Chapter of the Association for Computational Linguistics: Human Language
  Technologies, {NAACL-HLT} 2021, Online, June 6-11, 2021}, pages 380--396.
  Association for Computational Linguistics.

\bibitem[{Syed et~al.(2020)Syed, Baff, Kiesel, Khatib, Stein, and
  Potthast}]{syed:2020}
Shahbaz Syed, Roxanne~El Baff, Johannes Kiesel, Khalid~Al Khatib, Benno Stein,
  and Martin Potthast. 2020.
\newblock \href {https://doi.org/10.18653/v1/2020.coling-main.470} {News
  editorials: Towards summarizing long argumentative texts}.
\newblock In \emph{Proceedings of the 28th International Conference on
  Computational Linguistics, {COLING} 2020, Barcelona, Spain (Online), December
  8-13, 2020}, pages 5384--5396. International Committee on Computational
  Linguistics.

\bibitem[{Syed et~al.(2021)Syed, Khatib, Alshomary, Wachsmuth, and
  Potthast}]{syed:2021}
Shahbaz Syed, Khalid~Al Khatib, Milad Alshomary, Henning Wachsmuth, and Martin
  Potthast. 2021.
\newblock \href {http://arxiv.org/abs/2106.01064} {Generating informative
  conclusions for argumentative texts}.
\newblock \emph{CoRR}, abs/2106.01064.

\bibitem[{Toledo et~al.(2019)Toledo, Gretz, Cohen-Karlik, Friedman, Venezian,
  Lahav, Jacovi, Aharonov, and Slonim}]{toledo:2019}
Assaf Toledo, Shai Gretz, Edo Cohen-Karlik, Roni Friedman, Elad Venezian, Dan
  Lahav, Michal Jacovi, Ranit Aharonov, and Noam Slonim. 2019.
\newblock Automatic argument quality assessment-new datasets and methods.
\newblock In \emph{Proceedings of the 2019 Conference on Empirical Methods in
  Natural Language Processing and the 9th International Joint Conference on
  Natural Language Processing (EMNLP-IJCNLP)}, pages 5625--5635.

\bibitem[{Wang and Ling(2016)}]{wang:2016}
Lu~Wang and Wang Ling. 2016.
\newblock \href {https://doi.org/10.18653/v1/N16-1007} {Neural network-based
  abstract generation for opinions and arguments}.
\newblock In \emph{Proceedings of the 2016 Conference of the North {A}merican
  Chapter of the Association for Computational Linguistics: Human Language
  Technologies}, pages 47--57, San Diego, California. Association for
  Computational Linguistics.

\bibitem[{Zhang et~al.(2020)Zhang, Kishore, Wu, Weinberger, and
  Artzi}]{Zhang:2020}
Tianyi Zhang, Varsha Kishore, Felix Wu, Kilian~Q. Weinberger, and Yoav Artzi.
  2020.
\newblock \href {https://openreview.net/forum?id=SkeHuCVFDr} {Bertscore:
  Evaluating text generation with bert}.
\newblock In \emph{International Conference on Learning Representations}.

\end{thebibliography}
	\bibliographystyle{acl_natbib}

\end{document}